\documentclass[11pt]{article}
\usepackage[preprint]{acl}
\usepackage{times}
\usepackage{latexsym}
\usepackage[T1]{fontenc}
\usepackage[utf8]{inputenc}
\usepackage{microtype}
\usepackage{inconsolata}
\usepackage{graphicx}
\usepackage{amsmath}
\usepackage{amssymb}
\usepackage{booktabs}
\usepackage{multirow}
\usepackage{xcolor}
\usepackage{colortbl}
\usepackage{algorithm}
\usepackage{algorithmic}
\usepackage{enumitem}
\usepackage{subcaption}
\usepackage{placeins}
\graphicspath{{figures/}}

\newcommand{\pre}{Pre-CoT}
\newcommand{\intra}{Intra-CoT}
\newcommand{\post}{Post-CoT}

\title{Post-Training Shifts Confidence:
A Three-Stage Analysis of How SFT, RL, and OPD Shape CoT Calibration}

\author{
  Shuhao Li\textsuperscript{1,2}, 
  Guodong Du\textsuperscript{2}, 
  Anhao Zhao\textsuperscript{1,2}, 
  Wanyu Lin\textsuperscript{2}, 
  Tianyu Yuan\textsuperscript{1$\ast$}, 
  Xiaoyu Shen\textsuperscript{1}\thanks{Corresponding Author} \\
  \\
  \textsuperscript{1} Eastern Institute of Technology, Ningbo
  \quad
  \textsuperscript{2}The Hong Kong Polytechnic University
  \\
  \texttt{shuhao.li@connect.polyu.hk} \quad
  \texttt{tyyuan@eitech.edu.cn} \quad
  \texttt{xyshen@eitech.edu.cn} \\
}

\begin{document}
\maketitle

\begin{abstract}
Large language models have made strong reasoning gains through supervised fine-tuning, reinforcement learning, and on-policy distillation, yet these post-training methods are usually evaluated only by final-answer accuracy. We study how they reshape confidence during reasoning. We introduce a three-stage calibration framework that evaluates confidence before, during, and after chain-of-thought generation, corresponding to difficulty estimation, early termination, and answer aggregation. Through a controlled comparison on mathematical reasoning benchmarks, we find that OPD provides the most useful pre-reasoning confidence, SFT gives the strongest online signal for early stopping, and RL produces the most reliable trace-level signal for aggregation. We further show that confidence reliability is position-dependent: RL confidence becomes informative after a path-commitment phase, while OPD confidence is useful early but can become inversely calibrated later. Based on this observation, we propose PosConf, a position-aware confidence strategy that uses confidence only from reliable relative-position intervals. PosConf improves RL answer aggregation by 6.1 points over majority voting and consistently improves OPD early stopping under tight token budgets, with gains up to 4.3 points by avoiding its later inverse-calibration region, showing that \emph{confidence in reasoning models should be used both stage-wise and position-awarely}. Our code is available at \url{https://github.com/EIT-NLP/Post-Training-Calibration}.
\end{abstract}

\section{Introduction}
\label{sec:intro}

Large language models (LLMs) have achieved rapid progress on mathematical and logical reasoning tasks \citep{openai2024gpt4technicalreport,lewkowycz2022solvingquantitativereasoningproblems,yang2024qwen25mathtechnicalreportmathematical}.
This progress is largely driven by reasoning-oriented post-training techniques, such as supervised fine-tuning (SFT) on chain-of-thought traces \citep{wei2022chain,zelikman2022star,yu2024metamath}, on-policy distillation (OPD) from larger models \citep{Guo_2025,zhao2026poweropd,zhao2026decoupling}, and reinforcement learning (RL) with verifiable rewards \citep{schulman2017proximal,shao2024deepseekmath}. Despite their methodological differences, these methods are still commonly evaluated through single-attempt final-answer accuracy.

Final-answer accuracy, however, provides only a partial view of reasoning behavior. It measures whether a model eventually reaches the correct answer, but does not reveal whether the model can reliably estimate its own uncertainty during the reasoning process \citep{guo2017calibration,kadavath2022languagemodelsmostlyknow}. This distinction is especially important for inference-time compute. Before reasoning begins, confidence can serve as a signal of problem difficulty \citep{zhu2025llm,chen2025query}; during reasoning, it can help identify low-quality traces and terminate them early \citep{fu2025deep,wang2026taps}; and after reasoning, it can support the selection or aggregation of multiple sampled solutions \citep{wang2022self,brown2024large}. In these settings, well-calibrated confidence can improve not only final accuracy, but also the efficiency and reliability of reasoning \citep{snell2024scaling,wu2024inference}.

Existing research leaves two key questions unresolved. First, work on confidence estimation often treats calibration as an aggregate model property or a final-output signal, rather than studying where confidence is reliable within a chain-of-thought trace \citep{kuhn2023semantic,lin2022teaching,tian2023just,xiong2024can,geng2024survey,liu2025uncertainty}. Second, prior comparisons of reasoning-oriented post-training methods mainly focus on final-answer accuracy, leaving unclear how SFT, OPD, and RL affect confidence calibration and uncertainty estimation. As a result, we still lack a systematic understanding of how different post-training paradigms reshape confidence before, during, and after reasoning.

To bridge this gap, we systematically investigate how reasoning-oriented post-training affects confidence calibration across the entire reasoning pipeline. Rather than treating calibration as a monolithic property of the final output, we decompose the reasoning process into a stage-wise framework. This allows us to evaluate the utility of confidence signals for specific inference-time decisions through three targeted calibration tasks:
In the \pre{} stage, we evaluate whether confidence signals before chain-of-thought generation can predict problem difficulty.
In the \intra{} stage, we examine whether confidence signals during generation can identify reasoning traces that are unlikely to lead to correct answers and support early termination.
In the \post{} stage, we test whether confidence signals from complete reasoning traces can improve answer aggregation across multiple candidates.

To our knowledge, this is the first controlled study of how SFT, RL, and OPD reshape confidence calibration in reasoning models. We compare variants derived from the same Qwen2.5-7B-Instruct backbone and trained on the same problem mixture drawn from DeepScaler and SimpleRL, holding the model family and data source fixed so that calibration differences can be attributed directly to the post-training phase. The training setup is constructed to make the three paradigms comparable: SFT learns from DeepSeek-R1-Distill-Qwen-32B teacher traces retained only when the final answer is correct; OPD distills teacher token-level supervision onto student-generated rollouts; and RL optimizes verifiable outcome rewards. We then evaluate these models on diverse reasoning benchmarks.

This controlled setup reveals a sharper picture than final-answer accuracy alone. OPD provides the most reliable \pre{} signal for difficulty estimation, SFT yields the strongest \intra{} signal for early termination, and RL produces the best \post{} signal for answer aggregation. More surprisingly, we find that confidence reliability is strongly position-dependent. SFT maintains a relatively stable confidence gap between correct and incorrect traces, which explains why it is effective for online early stopping. RL confidence is initially uninformative, but becomes reliable after the model commits to a reasoning direction. OPD shows the opposite pattern: its early confidence is useful, but later confidence can become inversely calibrated, meaning that incorrect traces may appear more confident than correct ones. These patterns explain why naive confidence averaging over the full trace can be misleading: it mixes reliable and unreliable regions into one score.

Motivated by this observation, we introduce PosConf, a position-aware confidence strategy that uses confidence only from reliable relative-position intervals. PosConf is simple but effective because it matches confidence extraction to each model's calibration dynamics. For RL, it avoids making early decisions before confidence becomes meaningful, improving early stopping by up to about 4 points under matched token budgets and further raising confidence-based aggregation to a 6.1-point average gain over majority voting. For OPD, it avoids the later inverse-calibration region and instead relies on the earlier part of the trace, consistently improving low-budget early stopping across all benchmarks, with gains of more than 3 points on AIME-style problems. These results show that confidence in reasoning models should be used not only stage-wise, but also position-awarely.

\paragraph{Contributions.}
We make three main contributions.
First, we introduce a three-stage framework for reasoning models that connects confidence from \pre{}, \intra{} and \post{} to inference-time compute decisions.
Second, we provide a controlled comparison of SFT, RL, and OPD, showing that reasoning-oriented post-training induces distinct confidence shifts.
Third, we further show that confidence reliability is position-dependent and introduce PosConf, a position-aware strategy that improves confidence-based early stopping and aggregation.

\begin{figure*}[t]
\centering
\includegraphics[width=\textwidth,height=0.38\textheight,keepaspectratio]{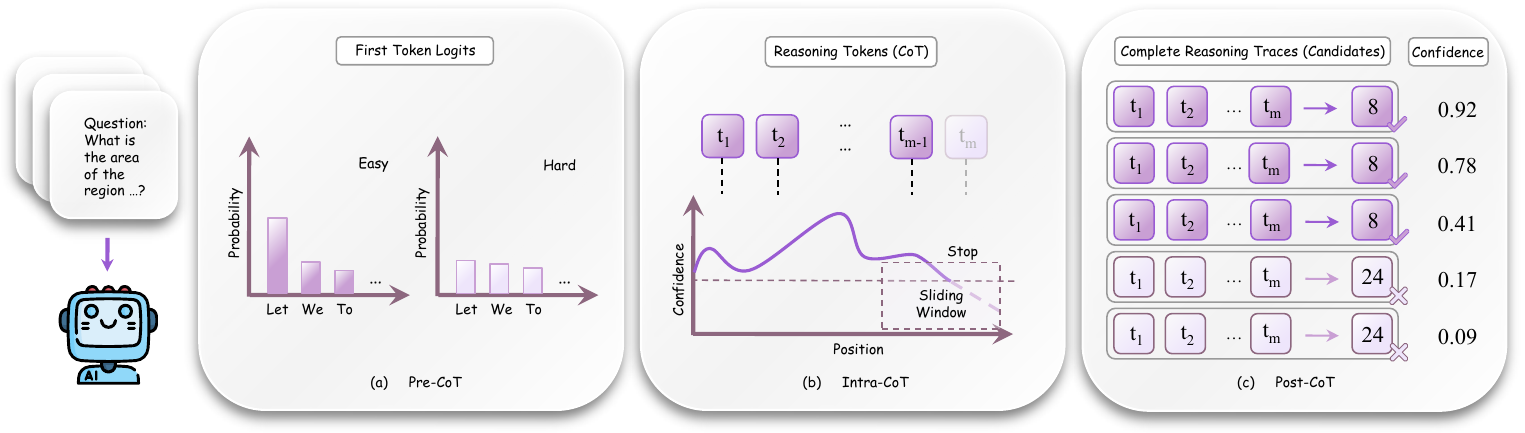}
\caption{
Overview of the three-stage reasoning calibration framework.
Confidence is evaluated before, during, and after chain-of-thought generation,
corresponding to difficulty estimation, early termination, and answer aggregation.
}
\label{fig:stagewise_reasoning_calibration}
\end{figure*}

\section{Confidence Calibration Framework}
\label{sec:framework}

We partition confidence by what decision it can affect: before generation for query-level routing, during generation for
online pruning, and after generation for candidate aggregation. This view yields three stages: \pre{}, \intra{}, and
\post{}.

Let $x$ denote an input problem with ground-truth answer $y$, and let $P_x$ denote the corresponding prompt. For each problem, the model samples a set of reasoning traces $\mathcal{T}_x=\{\tau_i\}_{i=1}^{N}$, where we use the simplified notation $\mathcal{T}$ and $\tau=(t_1,\dots,t_m)$ when the context is clear. Each trace $\tau_i$ produces an extracted answer $\mathrm{answer}(\tau_i)$ and receives a trace-level confidence score $C(\tau_i)$. As our foundational internal confidence signal, we rely on token-level predictive statistics. Given a prefix $s_\ell = (P_x, t_1, \dots, t_{\ell-1})$, let $p(v \mid s_\ell)$ be the model's next-token distribution over the vocabulary $\mathcal{V}$. Following \citet{fu2025deep}, we derive confidence from the top-$k$ candidate tokens, $\mathcal{V}_k(s_\ell)$. The token-level confidence at position $\ell$ is defined as:
\begin{equation}
    c_\ell = -\frac{1}{k} \sum_{v \in \mathcal{V}_k(s_\ell)} \log p(v \mid s_\ell).
\label{eq:token_conf}
\end{equation}
Intuitively, this score captures the concentration of the next-token distribution: a highly peaked distribution results in a larger confidence value, whereas a flatter, more uncertain distribution yields a lower value. We also use relative token position to compare confidence dynamics across
traces of different lengths. For a token at index $t$ in a trace of length $m$,
we define its relative token position as
\[
\rho_t = t/m \in [0,1].
\]
Thus, $\rho_t \approx 0$ corresponds to the beginning of the reasoning trace,
whereas $\rho_t \approx 1$ corresponds to the final-answer region.

\paragraph{The \pre{} Stage.} 
The \pre{} stage evaluates whether confidence \emph{before} generation begins can serve as a query-level signal to predict problem difficulty. We compute a query-level confidence score before any chain-of-thought tokens are generated by applying the token-level confidence formula to the prompt itself ($s_1=P_x$):
\begin{equation}
    C_{\mathrm{pre}}(x) = c_1(P_x) = -\frac{1}{k} \sum_{v \in \mathcal{V}_k(P_x)} \log p(v \mid P_x).
\label{eq:pre_conf}
\end{equation}
To proxy actual difficulty, we sample $\mathcal{T}_x$ and aggregate answers by
majority voting \citep{wang2022self}. A problem is ``solved'' if the
majority-voted answer $\hat{a}_x$ matches $y$, and ``unsolved''
otherwise. We then measure how well \pre{} confidence ranks these
outcomes using AUROC and PRR \citep{geifman2017selective}. Higher values
indicate that $C_{\mathrm{pre}}(x)$ captures model-specific difficulty before
reasoning starts.

\paragraph{The \intra{} Stage.}

The \intra{} stage evaluates whether confidence \emph{during} generation can serve as a step-level signal to detect unreliable reasoning paths. For a partial trace $\tau_{1:\ell}=(t_1,\dots,t_\ell)$, we compute token-level confidence scores $\{c_1,\dots,c_\ell\}$.  To reduce lexical and formatting noise in individual token probabilities, we adopt the group-confidence approach from \citet{fu2025deep} and smooth these values over a sliding window of size $w$. For any position $\ell \ge w$, the local window confidence is defined as:
\begin{equation}
    G_{\ell,w}(\tau) = \frac{1}{w} \sum_{r=\ell-w+1}^{\ell} c_r.
    \label{eq:group_conf}
\end{equation}
This metric acts as an online reliability signal. If $G_{\ell,w}(\tau)$ drops below a predefined threshold $\theta$, it suggests the trace has entered a highly uncertain segment, triggering early termination. These truncated traces are excluded from final answer aggregation. We evaluate this stage by sweeping the threshold $\theta$ to map the token-accuracy frontier.\footnote{The full Intra-CoT threshold-sweep procedure is provided in Algorithm~\ref{alg:intracot_sweep} in the appendix.}
A well-calibrated Intra-CoT signal should significantly reduce token usage without compromising accuracy.

\paragraph{The \post{} Stage.} 
The \post{} stage evaluates whether confidence \emph{after} generation concludes can serve as a trace-level signal to improve answer aggregation. Here, we use the mean token confidence as a trace-level score for each completed trace in $\mathcal{T}_x$. A reliable \post{} signal should assign higher scores to traces that yield correct answers. To evaluate this, we compare standard majority voting against confidence-filtered majority voting. While the standard approach selects the most frequent answer across all samples, our filtered approach first ranks traces by $C_{\mathrm{post}}(\tau_i)$, retains only the top $\eta$\%, and performs majority voting on this refined subset. We additionally report ``most-confident selection'', which simply extracts the answer from the single highest-confidence trace. Improved accuracy from these methods confirms that trace-level confidence successfully isolates reliable candidates.

\section{Experimental Setup}
\label{sec:setup}

\paragraph{Models.}
\label{sec:models}

We compare four models based on the Qwen2.5-7B-Instruct backbone
\citep{qwen2025qwen25technicalreport}. We refer to them as Qwen-Instruct, Qwen-SFT, Qwen-RL, and Qwen-OPD.
Qwen-Instruct is the original instruction-tuned backbone and serves as the
reference model without additional reasoning-oriented post-training. The other
three models are trained on the same mixture of DeepScaler \citep{deepscaler2025} and SimpleRL \citep{zeng2025simplerl} data,
but with different post-training paradigms.

Qwen-SFT is trained with standard cross-entropy on long
chain-of-thought traces generated by DeepSeek-R1-Distill-Qwen-32B \citep{Guo_2025}.
To construct high-quality supervised targets, we use rejection sampling and keep
only teacher responses whose final answers are correct. Qwen-RL is trained with group relative policy optimization
\citep{shao2024deepseekmath} using binary outcome rewards \citep{sheng2025hybridflow}. Qwen-OPD first samples rollouts from the student model and then distills
token-level supervision from DeepSeek-R1-Distill-Qwen-32B through dense KL
divergence \citep{yang2025qwen3technicalreport,jin2026entropy}. Holding the backbone
and data source fixed makes calibration differences more attributable to the
post-training objective. Detailed training configurations are provided in Appendix~\ref{app:training}.

\paragraph{Benchmarks.}
\label{sec:benchmarks}

We evaluate all models on four mathematical reasoning benchmarks with different
difficulty profiles. AIME 2024 \citep{aime24} and AIME 2025 \citep{aime25} are competition-level benchmarks consisting of challenging short-answer problems. AMC 2023 provides a broader set of moderate-difficulty contest problems, while MATH500 \citep{hendrycks2021measuring} offers a larger and more statistically stable evaluation set covering diverse mathematical settings.

\section{Results}
\label{sec:experiments}

We present results across three stages. Overall, we find that
reasoning-oriented post-training induces stage-dependent changes in confidence calibration.
In \pre{}, OPD provides the most informative pre-reasoning signal. In \intra{},
SFT yields the most effective token-level signal for early stopping. In \post{},
RL shows the largest benefit when confidence is used to filter completed reasoning traces. These results show that post-training does not
uniformly improve confidence calibration.

\paragraph{\pre{}: Difficulty Estimation}
\label{sec:precot_results}

\begin{table}[!t]
\caption{\pre{} calibration for problem difficulty estimation. Higher AUROC and
PRR indicate more reliable \pre{} confidence.}
\label{tab:precot}
\centering
\small
\setlength{\tabcolsep}{3pt}
\resizebox{\columnwidth}{!}{%
\begin{tabular}{@{}llccccc@{}}
\toprule
Model 
& Metric
& AIME 24
& AIME 25
& AMC 23
& MATH500
& Avg. \\
\midrule
\rowcolor[rgb]{ 1,  .953,  .949}
Qwen-Instruct
& AUROC & 0.618 & 0.729 & 0.415 & 0.613 & 0.594 \\
\rowcolor[rgb]{ 1,  .953,  .949}
& PRR   & 0.198 & 0.221 & -0.109 & 0.275 & 0.146 \\
\midrule
\rowcolor[rgb]{ .949,  .949,  1}
Qwen-SFT
& AUROC & 0.592 & 0.520 & 0.411 & 0.580 & 0.526 \\
\rowcolor[rgb]{ .949,  .949,  1}
& PRR   & 0.107 & 0.064 & -0.078 & 0.286 & 0.095 \\
\midrule
\rowcolor[rgb]{ .949,  1,  .949}
Qwen-RL
& AUROC & 0.464 & 0.382 & 0.180 & 0.255 & 0.320 \\
\rowcolor[rgb]{ .949,  1,  .949}
& PRR   & -0.151 & -0.203 & -0.702 & -0.854 & -0.478 \\
\midrule
\rowcolor[rgb]{  .949,  .949,  .949}
Qwen-OPD
& AUROC & 0.661 & 0.694 & 0.486 & 0.734 & \textbf{0.644} \\
\rowcolor[rgb]{  .949,  .949,  .949}
& PRR   & 0.324 & 0.375 & 0.070 & 0.611 & \textbf{0.345} \\
\bottomrule
\end{tabular}%
}
\end{table}

Table~\ref{tab:precot} shows that \pre{} confidence differs substantially across
model variants. Qwen-OPD achieves the strongest overall \pre{} calibration,
with the highest average AUROC of 0.644 and the highest average PRR of 0.345.
Its advantage is especially clear on MATH500, where it obtains an AUROC of
0.734 and a PRR of 0.611. Qwen-OPD also performs strongly on AIME 2024, where it
achieves the best AUROC and PRR among all models. On AIME 2025, Qwen-Instruct
achieves the highest AUROC, but Qwen-OPD still obtains the best PRR, indicating
that its confidence ranking remains more useful for selective prediction.

In contrast, Qwen-RL shows substantially weaker \pre{} calibration. Its AUROC
falls below 0.5 on all four benchmarks, and its PRR is negative across all
benchmarks. This suggests that Qwen-RL confidence before reasoning does not
reliably distinguish easy problems from hard ones. In some cases, it may even
mis-rank problem difficulty, making low-confidence rejection harmful rather than
helpful. Qwen-SFT also does not improve over Qwen-Instruct in the \pre{} stage,
suggesting that supervised fine-tuning on correct reasoning traces does not
necessarily preserve query-level difficulty perception.

Overall, Qwen-OPD provides the best-calibrated \pre{} signal,
whereas Qwen-RL shows a weaker ability to assess problem difficulty.

\paragraph{\intra{}: Early Stopping}
\label{sec:intracot_results}

\begin{figure*}[t]
\centering
\includegraphics[width=\textwidth]{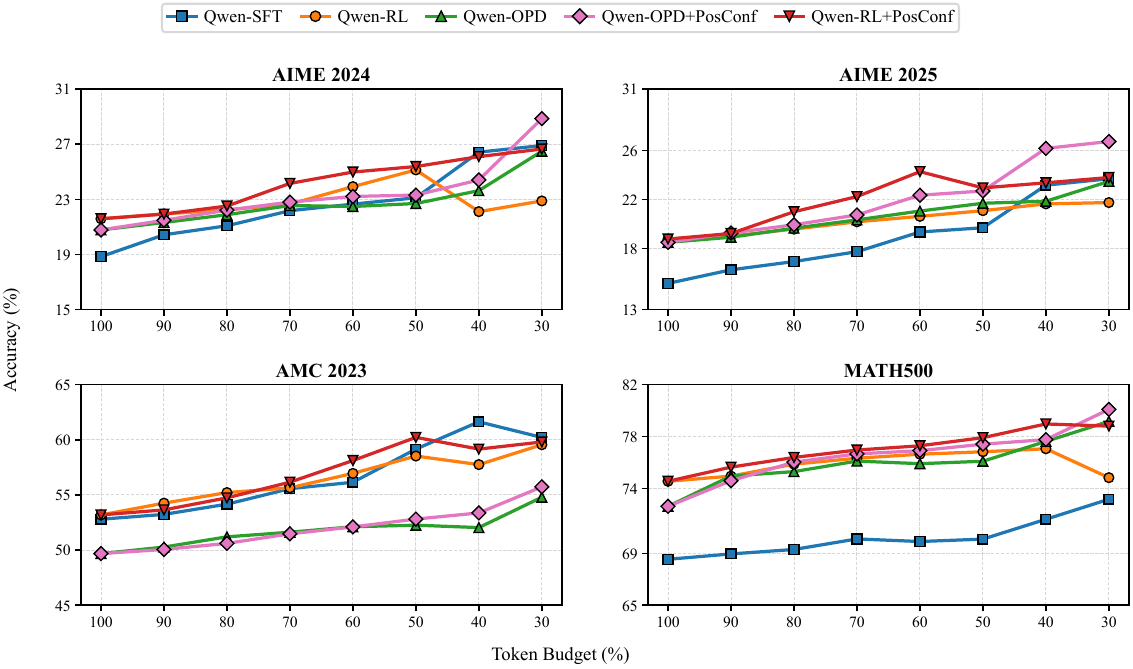}
\caption{Matched-budget \intra{} early-stopping frontiers under standard
confidence and PosConf. The x-axis denotes the retained token budget, from the
full online budget (100\%) to aggressive early stopping (30\%). PosConf uses
model-specific reliable relative-position intervals and improves low-budget frontiers when confidence
is position-dependent.}
\label{fig:intracot_perbenchmark}
\end{figure*}

Figure~\ref{fig:intracot_perbenchmark} shows the matched-budget \intra{}
frontiers, including both standard confidence-based early stopping and the
PosConf variants introduced in Section~\ref{sec:position_strategy}. The
100\% point corresponds to the full online budget, while moving right applies
more aggressive early stopping down to a 30\% retained budget.

Standard confidence-based filtering improves the accuracy--compute trade-off for
all post-trained models, but the improvement profiles differ across
training paradigms. Qwen-SFT gives the clearest \intra{} gains on the
competition-style benchmarks: from the 100\% to 30\% budget, accuracy rises from
18.84 to 26.90 on AIME 2024 and from 15.14 to 23.68 on AIME 2025. On AMC 2023,
it peaks at 61.63 under a 40\% budget, compared with 52.80 at the full budget.
The MATH500 gain is smaller but still positive, increasing from 68.54 to 73.17
at the 30\% budget.

Qwen-RL also benefits from filtering, but its gains are less stable under very
aggressive stopping. On AIME 2024, it improves from 21.59 to 25.13 at the 50\%
budget, but drops to 22.88 at the 30\% budget. On AIME 2025 and AMC 2023, the
30\% budget still improves accuracy from 18.76 to 21.74 and from 53.19 to
59.55, respectively. On MATH500, Qwen-RL peaks at 77.06 under a 40\% budget and
then falls back to 74.83 at 30\%.

Qwen-OPD exhibits a more aggressive low-budget frontier. At the 30\% budget, it
improves from 20.78 to 26.49 on AIME 2024, from 18.49 to 23.46 on AIME 2025,
and from 72.62 to 79.17 on MATH500. However, on AMC 2023 it remains below
Qwen-SFT and Qwen-RL even after filtering, reaching 54.79 at 30\% compared with
60.21 for Qwen-SFT and 59.55 for Qwen-RL. This indicates that OPD confidence can
preserve strong traces under tight budgets for some benchmarks.

\paragraph{\post{}: Answer Aggregation}
\label{sec:postcot_results}

\begin{table}[!t]
\caption{\post{} answer aggregation accuracy (\%). Confidence filtering benefits
Qwen-RL most, while naive confidence filtering
hurts Qwen-SFT and Qwen-OPD relative to majority voting.}
\label{tab:postcot}
\centering
\small
\renewcommand{\arraystretch}{0.88}
\setlength{\tabcolsep}{3pt}
\resizebox{\columnwidth}{!}{%
\begin{tabular}{@{}llccccc@{}}
\toprule
\textbf{Model} & \textbf{Aggregation} & \textbf{AIME 24} & \textbf{AIME 25} & \textbf{AMC 23} & \textbf{MATH500} & \textbf{Avg.} \\
\midrule
\rowcolor[rgb]{ 1,  .953,  .949}
\multirow[t]{4}{*}{Qwen-Instruct}
& Majority Vote & 20.0 & 18.7 & 65.0 & 81.1 & 46.2 \\
\rowcolor[rgb]{ 1,  .953,  .949}
& Most Confident & 20.0 & 14.7 & 54.0 & 80.4 & 42.3 \\
\rowcolor[rgb]{ 1,  .953,  .949}
& Top-5\% Conf. & 21.3 & 18.7 & 59.0 & 81.3 & 45.1 \\
\rowcolor[rgb]{ 1,  .953,  .949}
& Top-10\% Conf. & 23.3 & 17.3 & 64.0 & 81.0 & \textbf{46.4} \\
\midrule
\rowcolor[rgb]{ .949,  .949,  1}
\multirow[t]{7}{*}{Qwen-SFT}
& Majority Vote & 33.3 & 33.3 & 87.5 & 85.2 & \textbf{59.8} \\
\rowcolor[rgb]{ .949,  .949,  1}
& Most Confident & 26.7 & 19.3 & 64.5 & 78.4 & 47.2 \\
\rowcolor[rgb]{ .949,  .949,  1}
& Top-5\% Conf. & 26.0 & 32.7 & 71.0 & 84.6 & 53.6 \\
\rowcolor[rgb]{ .949,  .949,  1}
& Top-10\% Conf. & 28.7 & 31.3 & 74.5 & 83.0 & 54.4 \\
\rowcolor[rgb]{ .949,  .949,  1}
& PosConf-Most & 31.3 & 30.7 & 67.0 & 80.4 & 52.4 \\
\rowcolor[rgb]{ .949,  .949,  1}
& PosConf-Top-5\% & 26.7 & 28.7 & 71.5 & 85.9 & 53.2 \\
\rowcolor[rgb]{ .949,  .949,  1}
& PosConf-Top-10\% & 30.0 & 27.3 & 75.0 & 85.8 & 54.5 \\
\midrule
\rowcolor[rgb]{ .949,  1,  .949}
\multirow[t]{7}{*}{Qwen-RL}
& Majority Vote & 30.0 & 29.3 & 87.0 & 83.2 & 57.4 \\
\rowcolor[rgb]{ .949,  1,  .949}
& Most Confident & 33.3 & 28.0 & 85.5 & 80.8 & 56.9 \\
\rowcolor[rgb]{ .949,  1,  .949}
& Top-5\% Conf. & 36.7 & 30.0 & 87.5 & 86.2 & 60.1 \\
\rowcolor[rgb]{ .949,  1,  .949}
& Top-10\% Conf. & 36.7 & 40.7 & 87.0 & 86.7 & 62.8 \\
\rowcolor[rgb]{ .949,  1,  .949}
& PosConf-Most & 28.7 & 40.0 & 84.0 & 87.5 & 60.0 \\
\rowcolor[rgb]{ .949,  1,  .949}
& PosConf-Top-5\% & 34.0 & 44.0 & 88.5 & 87.6 & \textbf{63.5} \\
\rowcolor[rgb]{ .949,  1,  .949}
& PosConf-Top-10\% & 33.3 & 46.0 & 86.5 & 87.2 & 63.3 \\
\midrule
\rowcolor[rgb]{  .949,  .949,  .949}
\multirow[t]{7}{*}{Qwen-OPD}
& Majority Vote & 30.0 & 30.0 & 85.0 & 89.0 & \textbf{58.5} \\
\rowcolor[rgb]{  .949,  .949,  .949}
& Most Confident & 24.7 & 26.7 & 80.0 & 81.4 & 53.2 \\
\rowcolor[rgb]{  .949,  .949,  .949}
& Top-5\% Conf. & 21.3 & 25.3 & 78.5 & 82.8 & 52.0 \\
\rowcolor[rgb]{  .949,  .949,  .949}
& Top-10\% Conf. & 22.0 & 27.3 & 79.0 & 85.8 & 53.5 \\
\rowcolor[rgb]{  .949,  .949,  .949}
& PosConf-Most & 30.7 & 28.6 & 82.0 & 83.2 & 56.1 \\
\rowcolor[rgb]{  .949,  .949,  .949}
& PosConf-Top-5\% & 24.0 & 26.7 & 79.0 & 89.0 & 54.7 \\
\rowcolor[rgb]{  .949,  .949,  .949}
& PosConf-Top-10\% & 24.0 & 28.0 & 79.5 & 88.6 & 55.0 \\
\bottomrule
\end{tabular}%
}
\renewcommand{\arraystretch}{1.0}
\end{table}

Table~\ref{tab:postcot} evaluates whether \post{} confidence can improve answer
aggregation after complete reasoning traces are generated. Under standard
majority voting, Qwen-SFT achieves the highest average accuracy of 59.8\%,
followed by Qwen-OPD at 58.5\% and Qwen-RL at 57.4\%. This shows that majority
voting primarily reflects the base quality and diversity of sampled reasoning
traces, rather than the usefulness of trace-level confidence.

Confidence-based aggregation changes this ranking. Qwen-RL benefits the most
from confidence filtering: its Top-10\% confidence voting reaches an average
accuracy of 62.8\%, improving over its own majority-vote baseline by 5.4
percentage points. The gain is especially clear on AIME 2025, where Qwen-RL
improves from 29.3\% under majority voting to 40.7\% under Top-10\% confidence
voting. Qwen-RL also improves on AIME 2024 and MATH500, indicating that its
trace-level confidence is useful for selecting reliable completed traces.

In contrast, Qwen-SFT and Qwen-OPD do not benefit from the same confidence
filtering strategy. For Qwen-SFT, both Top-5\% and Top-10\% confidence voting
reduce average accuracy relative to majority voting. For Qwen-OPD, all
confidence-based strategies underperform its majority-vote baseline in average accuracy. This
suggests that stronger majority-vote accuracy does not necessarily imply better
\post{} calibration: a model may generate correct traces, but its trace-level
confidence may still fail to rank those traces reliably.

Overall, Qwen-RL provides the best-calibrated \post{} confidence signal for answer
aggregation. Unlike \pre{} and \intra{}, where Qwen-OPD and Qwen-SFT are more
effective, \post{} favors Qwen-RL because confidence filtering improves the
quality of the retained reasoning traces.

\section{Analysis}
\label{sec:position_analysis}

The results above suggest that post-training does not simply make confidence more or less reliable.
Instead, it changes where confidence is calibrated within the reasoning process.
To better understand these differences, we analyze confidence dynamics across token positions and trace-level distributions.
This analysis helps explain why the same confidence signal can support \pre{}, \intra{}, or \post{} differently across post-training methods.

\paragraph{Position-Dependent Confidence Separation}
\label{sec:position_dynamics}

\begin{figure*}[t]
\centering
\includegraphics[width=\textwidth]{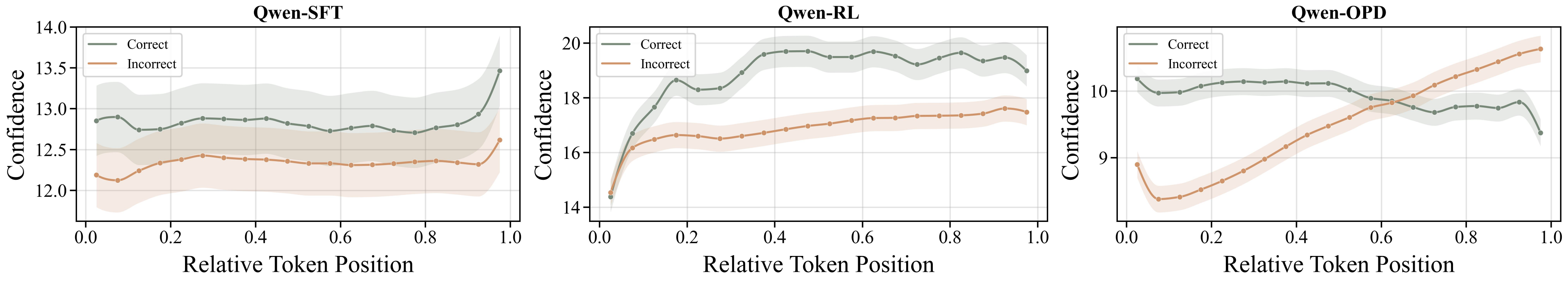}
\caption{Mean token-level confidence trajectories for correct and incorrect
traces on AIME 2024. Qwen-SFT preserves a stable confidence gap, Qwen-RL
becomes discriminative after an initial path-commitment phase, and Qwen-OPD
reverses direction later, with incorrect traces becoming more confident than
correct ones.}
\label{fig:token_trajectories}
\end{figure*}

Figure~\ref{fig:token_trajectories} shows the mean token-level confidence trajectories of correct and incorrect traces across relative token positions. Models with these post-training methods exhibit different patterns.

\begin{figure*}[t]
\centering
\includegraphics[width=\textwidth]{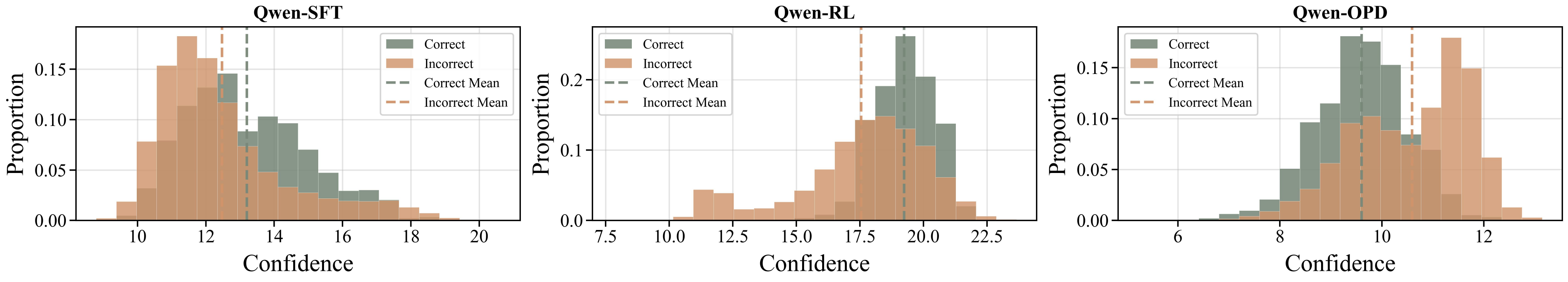}
\caption{Trace-level confidence distributions for correct and incorrect traces
on AIME 2024. Qwen-RL shows the clearest separation between correct and
incorrect traces, whereas Qwen-SFT has substantial overlap and Qwen-OPD exhibits
inverse calibration.}
\label{fig:confidence_distributions}
\end{figure*}

For Qwen-SFT, correct traces maintain higher confidence than incorrect traces across most of the reasoning process, with a sharper increase near the final-answer region.
This pattern is consistent with the SFT objective: token-level cross-entropy directly constrains the next-token distribution along teacher-generated reasoning traces.
As a result, local confidence fluctuations remain meaningful during generation, making SFT particularly suitable for \intra{}.
However, because SFT encourages sampled traces to follow similar teacher-like trajectories, its completed traces can have highly overlapping confidence distributions, as shown in Figure~\ref{fig:confidence_distributions}.
This limits the usefulness of trace-level confidence for \post{} answer aggregation.

Qwen-RL shows the opposite pattern.
At the beginning of reasoning, correct and incorrect traces are almost indistinguishable, which explains its weak \pre{} calibration.
After the model commits to a reasoning path, the two trajectories diverge rapidly.
The trace-level distribution in Figure~\ref{fig:confidence_distributions} further shows a clear separation: correct traces concentrate in a high-confidence region, while many incorrect traces form a lower-confidence tail.
This explains why RL benefits substantially from \post{} confidence filtering, even though its early-stage confidence is weak.

Qwen-OPD exhibits the most counterintuitive behavior.
It starts with well-calibrated early confidence, which is consistent with its strong \pre{} performance.
However, its confidence trajectories cross in the later part of reasoning: incorrect traces gradually become more confident and eventually surpass correct traces.
The trace-level distribution confirms this inverse calibration, where incorrect traces can receive higher confidence than correct ones.
This explains why \post{} confidence selection hurts Qwen-OPD.

\paragraph{PosConf: Confidence Extraction from Reliable Intervals}
\label{sec:position_strategy}

The trajectory analysis shows that confidence is not equally informative across relative token positions.
For SFT, confidence is informative across much of the trace and becomes especially discriminative near the final answer.
For RL, initial confidence is noisy, but the signal becomes reliable after a path-commitment phase.
For OPD, the initial part of the trace is informative, while later confidence becomes inversely calibrated after the confidence-transition point.
These observations explain why uniformly averaging confidence over an entire trace, or using a fixed tail window for all models, can be suboptimal.

To test whether confidence can be recovered from reliable relative-position intervals,
we consider a position-aware confidence score, which we call \textbf{PosConf}.
For a completed trace $\tau=(t_1,\ldots,t_m)$, we define
\begin{equation}
    C_w(\tau)
    =
    \frac{\sum_{t=1}^{m} w(t/m)c_t}
    {\sum_{t=1}^{m} w(t/m)},
    \label{eq:position_weighted_conf}
\end{equation}
where $c_t$ is the token-level confidence at position $t$, and $w(\cdot)$ is a position-dependent weighting function.
This formulation extracts trace-level confidence from relative-position intervals
where the signal is both discriminative and correctly oriented.
In our analysis, SFT and RL use tail confidence, while OPD uses initial confidence.

The same principle applies to \intra{}.
A standard early-stopping rule applies the same threshold throughout generation.
This can be misleading when confidence estimates are miscalibrated in certain positions. For RL, applying the threshold too early may stop traces before confidence becomes discriminative. For OPD, applying the threshold after the transition point may preferentially stop correct traces, because incorrect traces have become more confident. Therefore, early stopping should also be position-dependent.

Early stopping is activated only when the relative token position
$\rho_t=t/m$ lies within an active interval $I$ where confidence is well-calibrated.
Once this condition is satisfied, the trace is stopped if its windowed confidence $\mathrm{WinConf}_w(\tau,t)$ falls below a threshold $\theta$.
We parameterize the reliable intervals as two relative-position intervals:
$\mathcal{I}_{\mathrm{intra}}=[\alpha_{\mathrm{intra}},\beta_{\mathrm{intra}}]$
for \intra{}, and
$\mathcal{I}_{\mathrm{post}}=[\alpha_{\mathrm{post}},\beta_{\mathrm{post}}]$
for \post{}.
Table~\ref{tab:position_strategies} summarizes the PosConf parameters derived from AIME 2024 trajectory analysis.

\begin{table}[t]
\caption{Model-specific PosConf intervals derived from AIME 2024 trajectory analysis.
Each interval is defined over the relative token position $\rho=t/m \in [0,1]$.}
\label{tab:position_strategies}
\centering
\small
\begin{tabular*}{\columnwidth}{@{\extracolsep{\fill}}lcc@{}}
\toprule
Model & Intra-CoT interval & Post-CoT interval \\
\midrule
Qwen-SFT  & $[0, 1]$ & $[0.6, 1]$ \\
Qwen-RL   & $[0.2, 1]$ & $[0.6, 1]$ \\
Qwen-OPD  & $[0, 0.6]$ & $[0, 0.6]$ \\
\bottomrule
\end{tabular*}
\end{table}

The PosConf rows in Table~\ref{tab:postcot} show the resulting \post{} aggregation accuracy.
For Qwen-RL, PosConf further improves confidence-based aggregation, with PosConf-Top-5\% confidence voting reaching the strongest average performance.
For Qwen-OPD, extracting confidence from the pre-transition region mitigates inverse calibration.
For Qwen-SFT, PosConf scoring remains less effective than majority voting, suggesting that its confidence separation is too weak for aggressive filtering.

PosConf also improves \intra{} early stopping, as shown
in Figure~\ref{fig:intracot_perbenchmark}. For Qwen-RL, the 15\% skip period avoids premature stopping in the initial phase where correct and incorrect traces are difficult to distinguish from each other. This is most visible when standard early stopping becomes too aggressive: at the 40\% budget, PosConf improves AIME 2024 from 22.11 to 26.10 and MATH500 from 77.06 to 78.97; at the 60\% budget, it improves AIME 2025 from 20.61 to 24.26. On AMC 2023, the best matched-budget value rises from 59.55 to 60.22 with PosConf.

For Qwen-OPD, activating early stopping only before the confidence-transition region consistently improves low-budget performance. At the 30\% budget, PosConf raises accuracy from 26.49 to 28.85 on AIME 2024, from 23.46 to 26.72 on AIME 2025, from 54.79 to 55.73 on AMC 2023, and from 79.17 to 80.09 on MATH500. The gain is especially clear on AIME 2025, where PosConf adds 4.32 points at the 40\%
budget and 3.26 points at the 30\% budget.

Overall, these results show that confidence in reasoning models is a dynamic
and position-dependent signal. A model may provide reliable confidence early in
generation but become less reliable later, or provide uninformative confidence
early while becoming useful after committing to a reasoning path. PosConf offers
a practical way to align the use of confidence with where it is most reliable
within the reasoning trace.

\section{Discussion}
\label{sec:discussion}

\paragraph{Calibration is stage-dependent.}
Our results suggest that confidence calibration in reasoning models should not
be treated as a single global property of a model. A model can provide useful
confidence before reasoning begins, but unreliable confidence after a full trace
is generated; conversely, a model can be poorly calibrated at the query level
while still producing useful trace-level confidence for answer aggregation.
This distinction is important because confidence is used for different
inference-time decisions at different stages. Our findings show
that these uses are not interchangeable. The same confidence signal can be
helpful for one decision and misleading for another.

This stage-dependent view also changes how post-training should be evaluated.
Reasoning-oriented post-training is usually judged by final-answer accuracy, but
final accuracy alone does not determine whether a model's confidence is reliable.
For example, majority voting accuracy mainly reflects the quality and diversity
of sampled traces, whereas confidence-filtered aggregation tests whether the
model can identify which completed traces are reliable. Similarly, early
stopping requires confidence to be locally meaningful during generation, not
only correlated with the final answer after the trace is complete. Therefore,
evaluating calibration only at the final output can obscure important differences
among post-training methods and variants.

\paragraph{Post-training objectives shape where confidence is reliable.}
The different behaviors of SFT, RL, and OPD suggest that confidence dynamics are
closely tied to the training objective. SFT directly optimizes token-level
likelihood on correct teacher-generated reasoning traces. This objective appears
to preserve local confidence information throughout generation, which explains
why SFT is particularly effective for \intra{} early stopping. However, because
SFT encourages traces to follow similar teacher-like patterns, completed traces
may have overlapping confidence distributions, limiting the usefulness of
trace-level confidence for \post{} aggregation.

RL produces a different pattern. Since RL is driven by outcome rewards rather
than dense token-level supervision, its confidence before reasoning begins is a
weak indicator of whether the problem will be solved. This helps explain the
poor \pre{} calibration of the RL model. However, once the model commits to a
reasoning path, correct and incorrect traces become more separable in confidence
space. As a result, RL provides the strongest \post{} signal: confidence
filtering can identify reliable completed traces and improve answer aggregation.

As for OPD, its pre-reasoning confidence is relatively
informative, suggesting that distillation can preserve useful query-level
uncertainty. However, later in the reasoning trace, OPD confidence can become
misleading: incorrect traces may become more confident than correct traces.
This inverse calibration explains why naive confidence filtering hurts OPD in
\post{} aggregation. More broadly, this shows that distillation can transfer or
amplify confidence patterns that are useful in one region of the reasoning
process but unreliable in another.

\section{Conclusion}

We introduced a three-stage framework for evaluating confidence calibration in reasoning models before, during, and after chain-of-thought generation. Our results show that post-training induces stage-dependent confidence shifts: OPD is most effective for Pre-CoT difficulty estimation, SFT for Intra-CoT early stopping, and RL for Post-CoT answer aggregation.

We further find that confidence reliability is position-dependent within reasoning traces. Motivated by this, PosConf extracts confidence from reliable relative-position intervals, improving confidence-based early stopping and aggregation. These findings highlight the need to evaluate confidence not as a global model property, but as a stage-specific signal tied to different decisions.

\section*{Limitations}
\label{sec:limitations}

Our study focuses on mathematical reasoning benchmarks and models derived from a
single Qwen2.5-7B-Instruct backbone. This controlled design helps isolate the
effect of post-training paradigms, but future work should test whether the same
stage-dependent calibration patterns hold across larger model scales, different
model families, and non-mathematical reasoning tasks. In addition, we focus on
internal confidence derived from token probabilities. Other uncertainty signals,
such as entropy, probability margins, consistency across samples, or hybrid
confidence estimators, may capture complementary aspects of reasoning
reliability.

\FloatBarrier

\bibliography{custom}

\appendix

\section{Related Work}
\label{sec:related}

\paragraph{Chain-of-Thought Reasoning.}
Chain-of-thought (CoT) prompting \citep{wei2022chain} demonstrated that providing step-by-step reasoning examples enables large language models to solve complex problems.
Zero-shot variants \citep{kojima2022large} and scratchpad-based approaches \citep{nye2021workscratchpadsintermediatecomputation} further showed that reasoning capabilities can be elicited without task-specific exemplars.
Self-consistency \citep{wang2022self} improved reliability by sampling multiple reasoning paths and aggregating answers through majority voting, establishing a connection between sampling diversity and answer quality.
More structured approaches, such as Tree of Thoughts \citep{yao2023tree}, extend chain-of-thought reasoning by exploring multiple reasoning branches.
These works focus on how to generate or structure reasoning traces; our work complements them by studying how confidence signals along these traces can be used for inference-time decisions.

\paragraph{Reasoning-Oriented Post-Training.}
Several post-training paradigms have been developed to improve reasoning in LLMs.
Supervised fine-tuning on chain-of-thought traces, using data generated by stronger models or bootstrapped from the model itself \citep{zelikman2022star,yu2024metamath,singh2023beyond,chen2025unveiling}, directly trains models to produce step-by-step solutions.
Reinforcement learning with outcome-based rewards \citep{schulman2017proximal,shao2024deepseekmath,Guo_2025,deepscaler2025,zeng2025simplerl} optimizes reasoning through trial and error, rewarding correct final answers without requiring step-level supervision.
On-policy distillation \citep{hinton2015distilling,yang2025qwen3technicalreport,jin2026entropy,zhao2026self} combines elements of both by distilling token-level supervision from a teacher model onto student-generated rollouts.
Process reward models \citep{lightman2024let,uesato2022solving,wang2024math} provide step-level feedback but require additional verifier training.
Prior comparisons of these paradigms have focused on final-answer accuracy \citep{hosseini2024v,singh2023beyond}.
Our work differs by comparing how each paradigm reshapes the model's internal confidence signal across different stages of reasoning.

\paragraph{Calibration and Confidence Estimation in LLMs.}
Calibration measures whether a model's expressed confidence aligns with its actual accuracy \citep{guo2017calibration,naeini2015obtaining,platt1999probabilistic}.
Classical post-hoc approaches such as temperature scaling \citep{guo2017calibration} and Platt scaling \citep{platt1999probabilistic} were developed for classification tasks with well-defined output probabilities, while more recent work has revisited their effectiveness in modern deep networks \citep{minderer2021revisiting}.
For LLMs, confidence estimation has been studied through both internal signals and verbalized outputs.
Internal approaches extract confidence from token logits, hidden states, or predictive entropy \citep{kadavath2022languagemodelsmostlyknow,jiang2021can,kuhn2023semantic}.
Verbalized approaches elicit confidence through prompting strategies \citep{lin2022teaching,tian2023just,xiong2024can}.
\citet{desai2020calibration} showed that pre-trained transformers exhibit miscalibration that persists after fine-tuning, and \citet{joshi2025calibration} studied how calibration evolves across model layers.
Recent surveys \citep{geng2024survey,liu2025uncertainty} provide comprehensive overviews of this area.
\citet{zhu2025llm} showed that hidden representations before generation can estimate question difficulty.
\citet{fu2025deep} introduced group-level token confidence for reasoning models.
Our work extends calibration analysis to reasoning models by introducing a stage-wise framework that connects confidence to specific inference-time decisions, rather than measuring calibration as a single aggregate property.

\paragraph{Test-Time Compute Scaling.}
Recent work has shown that allocating more compute at inference time can substantially improve reasoning performance.
\citet{snell2024scaling} demonstrated that scaling test-time compute through repeated sampling and selection can be more effective than scaling model parameters.
\citet{brown2024large} showed that increasing the number of sampled solutions follows predictable scaling laws for problem coverage.
\citet{wu2024inference} analyzed compute-optimal inference strategies for problem-solving.
These approaches rely on mechanisms such as majority voting \citep{wang2022self,cobbe2021trainingverifierssolvemath}, best-of-$N$ selection with reward models \citep{lightman2024let,stiennon2020learning}, and adaptive compute allocation.
Entropy-based signals have also been explored for guiding RL training \citep{wang2025harnessing} and identifying critical reasoning tokens \citep{wang2026beyond}.
Our three-stage framework provides a structured view of when internal confidence supports different test-time strategies: \pre{} for difficulty-based routing, \intra{} for early termination, and \post{} for confidence-based answer aggregation.

\section{Training Details}
\label{app:training}

\paragraph{Backbone and training data.}
All model variants are derived from Qwen2.5-7B-Instruct. Qwen-Instruct is the
unmodified instruction-tuned backbone. Qwen-SFT, Qwen-RL, and Qwen-OPD are trained
on the same reasoning-oriented data mixture based on DeepScaler and SimpleRL, so
that differences in confidence behavior are primarily attributable to the
post-training objective rather than model family or data domain.

\paragraph{Supervised fine-tuning.}
Qwen-SFT is trained with full-parameter supervised fine-tuning using LlamaFactory.
The supervised targets are long chain-of-thought traces generated by
DeepSeek-R1-Distill-Qwen-32B and retained only when the extracted final answer is
correct. We use a cutoff length of 16,384 tokens, learning rate $1\times10^{-6}$,
two epochs, per-device batch size 1, gradient accumulation 8, cosine learning
rate schedule, warmup ratio 0.1, maximum gradient norm 1.0 and bf16 training.

\paragraph{Reinforcement learning.}
Qwen-RL is trained with group relative policy optimization using verifiable
answer rewards. The RL run uses maximum response length 16,384, train batch size
16, rollout count 8, KL coefficient $1\times10^{-4}$, entropy coefficient
$1\times10^{-3}$, rollout tensor parallel size 2, and 8 H20 GPUs. The reward is based
on whether the extracted final answer matches the ground truth.

\paragraph{On-policy distillation.}
Qwen-OPD uses on-policy distillation from DeepSeek-R1-Distill-Qwen-32B. The
student first generates its own rollouts, and the teacher provides dense
token-level supervision on those rollouts. We use the GOLD-style on-policy setup
with $\lambda=1.0$ and universal logit distillation to support teacher--student
tokenizer differences.

\section{Inference and Evaluation Protocol}
\label{app:inference}

\paragraph{Generation setup.}
For each model and benchmark problem, we generate multiple reasoning traces with
temperature 0.6, top-$p$ 0.95, and maximum generation length 32,768 tokens. The
generation scripts use an offline trace budget of 320 traces per problem. Unless
otherwise stated, aggregation experiments evaluate a fixed budget of 256 traces
per problem such that majority voting and confidence-filtered voting are compared
under the same sampled-trace budget. We use top-$k=20$ next-token log-probabilities for all confidence computations.

\paragraph{Repeated evaluation.}
To reduce variance from stochastic generation, we repeat each experiment five
times with different random seeds. Each repetition independently generates a new
set of reasoning traces and runs the same extraction, aggregation, and evaluation
pipeline. Unless otherwise stated, we report the mean performance over the five
repetitions.

\paragraph{Prompt template.}
All models use the same chat-style prompt. The user message appends the
instruction: ``Please reason step by step, and put your final answer within
\texttt{\textbackslash boxed\{\}}.'' This keeps answer formatting consistent
across models and benchmarks.

\paragraph{Answer extraction.}
We extract the final answer from each generated trace and compare it with the
benchmark ground truth. When mathematical equivalence checking is available, we
use it before falling back to normalized exact match. Invalid, missing, or
unparseable answers are counted as incorrect. The same extraction and evaluation
pipeline is used for all models.

\paragraph{\pre{} labels.}
For each model--problem pair, we label a problem as solved if the majority-voted
answer over sampled traces matches the ground truth, and unsolved otherwise.
This makes difficulty labels model-specific: a problem may be easy for one
post-training method and hard for another. We then evaluate whether first-token
confidence ranks solved problems above unsolved problems using AUROC and PRR.

\paragraph{\intra{} sweep.}
For online early stopping, we compute token confidence from the top-$k$
next-token probabilities and smooth it with a sliding window. A trace is stopped
when the windowed confidence falls below a threshold. Thresholds are derived from
warmup traces and increased to produce matched-budget token--accuracy frontiers. For each problem, we construct $M=50$ thresholds between the minimum warmup
window confidence and the 100 percentile of warmup window confidence. We then
select the operating points whose observed retained-token ratios are closest to token budgets in Figure~\ref{fig:intracot_perbenchmark}.
In each sweep, we use 32 warmup traces to estimate the minimum confidence and
adaptive window size, evaluate 256 online samples, and repeat the experiment
with 5 independent resamples unless otherwise specified.
To avoid future-information leakage, the online \intra{} variant does not
normalize token positions by the final length of the current trace. Instead, it
estimates an expected trace length $\hat{m}_x$ from warmup traces and uses the
online relative position $\hat{\rho}_t=\min\{t/\hat{m}_x,1\}$ when activating
position-restricted stopping rules. We set the adaptive divisor to $D=10$, so that the window size is
$w_x=\max\{1,\mathrm{round}(\bar L_x/D)\}$, where $\bar L_x$ is the mean
warmup trace length for problem $x$.

\begin{algorithm}[t]
\caption{\intra{} sweep}
\label{alg:intracot_sweep}
\begin{algorithmic}[1]
\STATE \textbf{Input:} model $p_\phi$, problem $x$, ground-truth answer $y_x$, reliable interval $\mathcal{I}_\phi$, warmup size $W=32$, online sample size $N=256$, sweep count $M$, adaptive divisor $D$, repetitions $R=5$
\FOR{each repetition $r=1,\ldots,R$}
\STATE Generate warmup traces $\mathcal{W}_x$ with $|\mathcal{W}_x|=W$
\STATE Compute each warmup trace length $L(\tau)$ and estimate $\hat{m}_x=\mathrm{median}_{\tau\in\mathcal{W}_x}L(\tau)$
\STATE Set the adaptive window size
$w_x=\max\{1,\mathrm{round}(\frac{1}{D}\cdot \frac{1}{|\mathcal{W}_x|}\sum_{\tau\in\mathcal{W}_x} L(\tau))\}$
\STATE For every warmup trace $\tau$, compute window confidence
$G_{t,w_x}(\tau)=\frac{1}{w_x}\sum_{q=t-w_x+1}^{t} c_q$
\STATE Define $\hat{\rho}_t=\min\{t/\hat{m}_x,1\}$ and compute $m_{\mathcal{I}}(\tau)=\min_{t:\hat{\rho}_t\in\mathcal{I}_\phi}G_{t,w_x}(\tau)$
\STATE Set $\theta_1=\min_{\tau\in\mathcal{W}_x}m_{\mathcal{I}}(\tau)$ and construct an increasing sweep $\Theta_x=\{\theta_1,\ldots,\theta_M\}$
\FOR{$\theta\in\Theta_x$}
\STATE Generate $N$ online samples. At token $t$, compute $G_{t,w_x}$ and $\hat{\rho}_t=\min\{t/\hat{m}_x,1\}$
\STATE Stop a sample only if $\hat{\rho}_t\in\mathcal{I}_\phi$ and $G_{t,w_x}<\theta$; otherwise continue generation
\STATE Record retained token ratio, surviving-path accuracy, valid-answer accuracy, and majority-vote accuracy
\ENDFOR
\ENDFOR
\STATE Aggregate the resulting token--accuracy points across problems and repetitions
\end{algorithmic}
\end{algorithm}

\paragraph{\post{} aggregation.}
For answer aggregation, majority voting uses all sampled valid answers.
Confidence filtering first ranks traces by trace-level confidence, keeps the
top 5\% or top 10\%, and then performs weighted majority voting over the retained
answers. We also report most-confident selection, which returns the answer from
the single highest-confidence trace. We repeat the post-CoT aggregation
experiment with 5 independent subsamples and report the averaged results unless
otherwise specified.

\section{Additional Metric Definitions}
\label{app:metrics}

\paragraph{AUROC.}
Let $s_i$ be a confidence score where larger values indicate higher confidence.
Let $\mathcal{P}$ and $\mathcal{N}$ denote correct and incorrect traces,
respectively. We compute trace-level AUROC as
\begin{equation}
\begin{aligned}
\mathrm{AUROC}
&= \frac{1}{|\mathcal{P}||\mathcal{N}|}
\sum_{i\in\mathcal{P}}\sum_{j\in\mathcal{N}} a_{ij}, \\
a_{ij}
&= \mathbf{1}\{s_i>s_j\}
+\frac{1}{2}\mathbf{1}\{s_i=s_j\}.
\end{aligned}
\end{equation}
An AUROC above 0.5 means confidence tends to rank correct traces above incorrect
traces; an AUROC below 0.5 indicates inverse calibration.

\paragraph{Prediction-Rejection Ratio.}
For selective prediction, sort examples by decreasing confidence and let
$\mathrm{Acc}(r)$ be the accuracy of the top $r$ retained examples. We define the
ranked area under the selection curve as
\begin{equation}
\mathrm{AUC}_{\mathrm{rank}}
= \frac{1}{n}\sum_{r=1}^{n}\mathrm{Acc}(r).
\end{equation}
Let $\mathrm{AUC}_{\mathrm{rand}}$ be the expected area under random rejection
and $\mathrm{AUC}_{\mathrm{oracle}}$ be the area obtained by ranking all correct
examples before incorrect examples. The prediction-rejection ratio is
\begin{equation}
\mathrm{PRR}
= \frac{
\mathrm{AUC}_{\mathrm{rank}}-\mathrm{AUC}_{\mathrm{rand}}
}{
\mathrm{AUC}_{\mathrm{oracle}}-\mathrm{AUC}_{\mathrm{rand}}
}.
\end{equation}
PRR is therefore a normalized area for the selective prediction curve. Negative
PRR indicates that confidence ranking is worse than random rejection.

\paragraph{Supplementary figures.}
In addition to the AIME 2024 trajectory and histogram figures in the main text,
we report the same diagnostics for AIME 2025, AMC 2023, and MATH500. Together,
these figures show that the position-dependent pattern is consistent across all
benchmarks.

\begin{figure*}[p]
\centering
\captionsetup[subfigure]{font=small,skip=1pt}

\begin{subfigure}{0.96\textwidth}
\centering
\includegraphics[width=\linewidth,height=0.115\textheight,keepaspectratio]{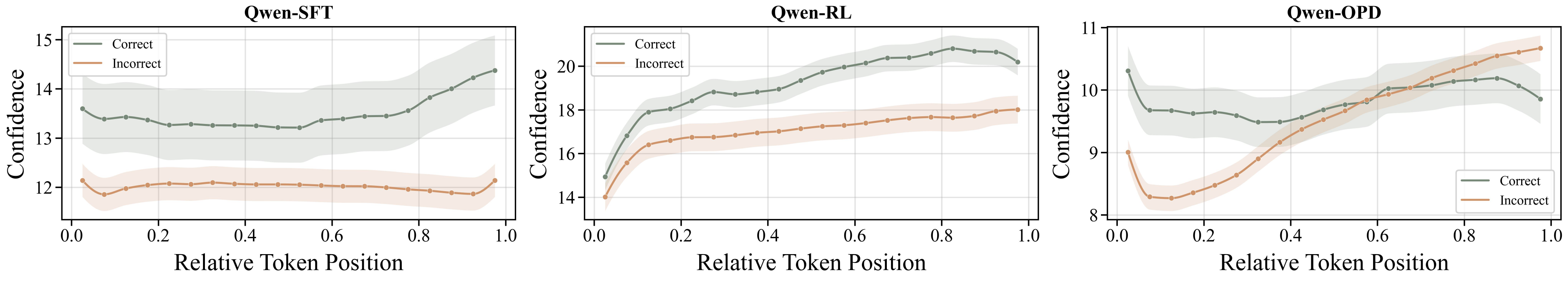}
\caption{AIME 2025: token trajectory}
\end{subfigure}

\vspace{0.25em}

\begin{subfigure}{0.96\textwidth}
\centering
\includegraphics[width=\linewidth,height=0.115\textheight,keepaspectratio]{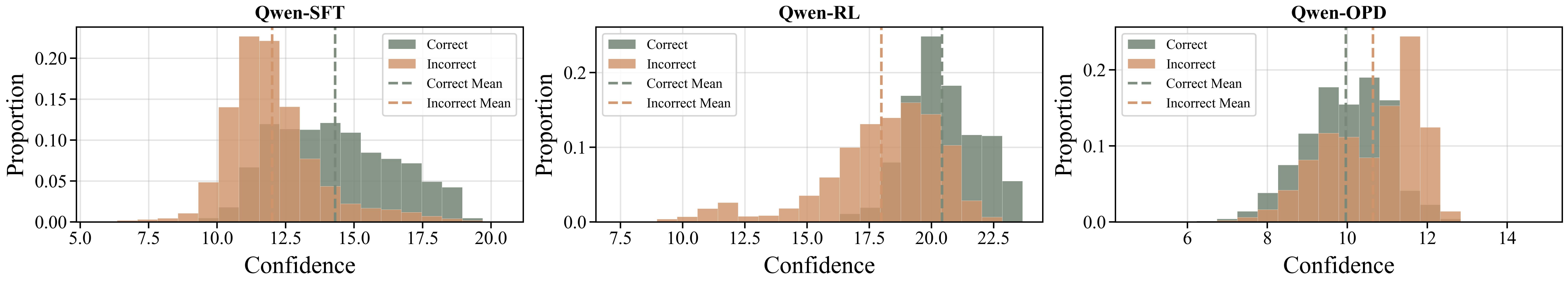}
\caption{AIME 2025: confidence histogram}
\end{subfigure}

\vspace{0.25em}

\begin{subfigure}{0.96\textwidth}
\centering
\includegraphics[width=\linewidth,height=0.115\textheight,keepaspectratio]{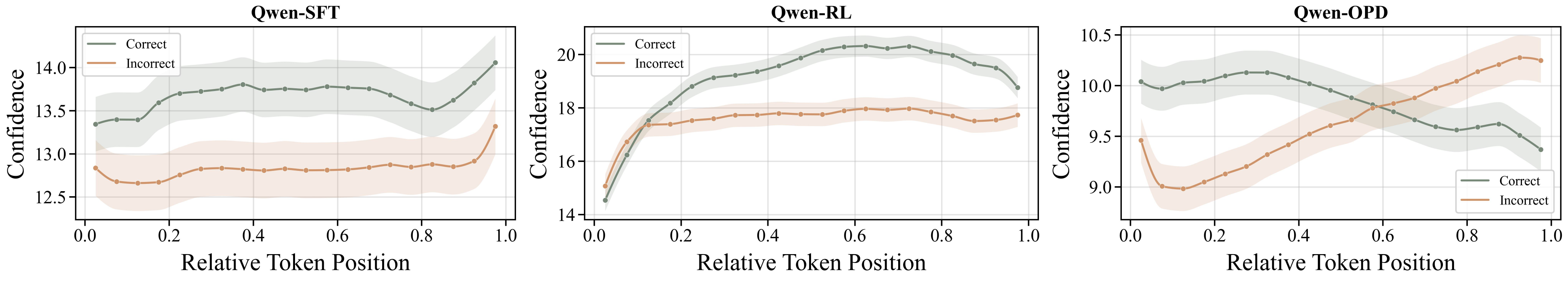}
\caption{AMC 2023: token trajectory}
\end{subfigure}

\vspace{0.25em}

\begin{subfigure}{0.96\textwidth}
\centering
\includegraphics[width=\linewidth,height=0.115\textheight,keepaspectratio]{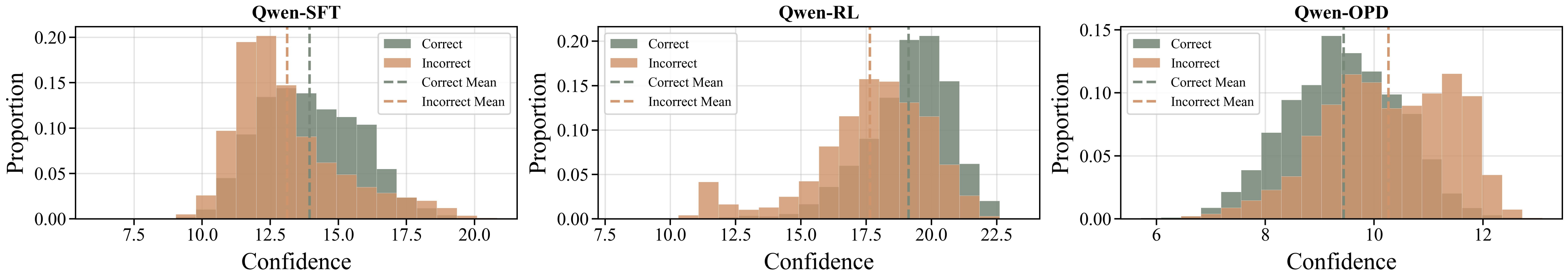}
\caption{AMC 2023: confidence histogram}
\end{subfigure}

\vspace{0.25em}

\begin{subfigure}{0.96\textwidth}
\centering
\includegraphics[width=\linewidth,height=0.115\textheight,keepaspectratio]{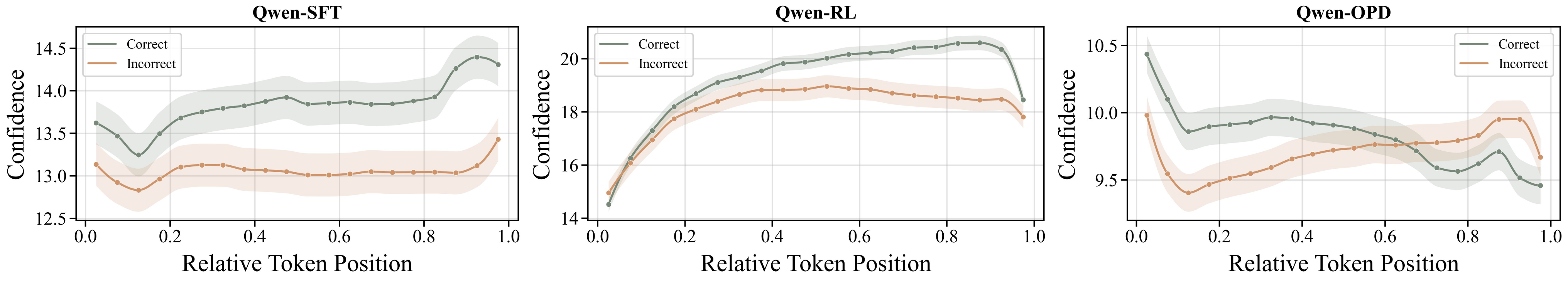}
\caption{MATH500: token trajectory}
\end{subfigure}

\vspace{0.25em}

\begin{subfigure}{0.96\textwidth}
\centering
\includegraphics[width=\linewidth,height=0.115\textheight,keepaspectratio]{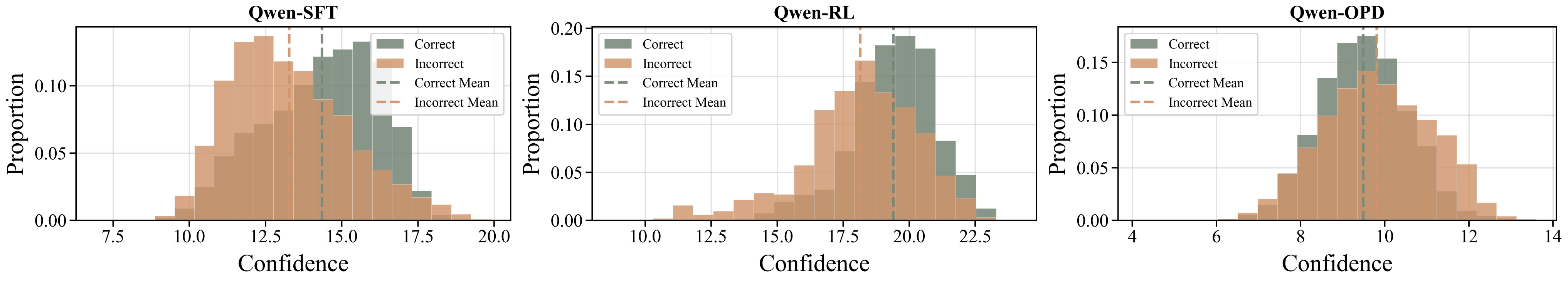}
\caption{MATH500: confidence histogram}
\end{subfigure}

\caption{
Supplementary confidence diagnostics on AIME 2025, AMC 2023, and MATH500. These plots test whether the
stage-dependent calibration patterns observed on AIME 2024 persist across benchmarks with different difficulty.
}
\label{fig:supp_confidence_diagnostics}
\end{figure*}

\end{document}